%% file: lmrl2025_workshop.tex
\pdfoutput=1
\documentclass{article} 
\usepackage{lmrl2025_workshop,times}

\usepackage{booktabs}
\usepackage{graphicx}
\usepackage{subcaption}
\graphicspath{{imgs/}}

\input{math_commands.tex}

\usepackage{hyperref}
\usepackage{url}
\RequirePackage{xcolor}

\newif\ifshowrevisions
\showrevisionsfalse

\DeclareRobustCommand{\changed}[1]{%
  \ifshowrevisions{\color{blue}#1}\else#1\fi%
}

\title{Are Vision Foundation Models \\ Foundational for Electron Microscopy \\ Image Segmentation?}

\author{Caterina Fuster-Barcel\'o \\
Department of Molecular Life Sciences,\\ Universit\"at Z\"urich\\
Z\"urich, Switzerland \\
\texttt{caterina.fusterbarcelo@mls.uzh.ch} \\
\And
Virginie Uhlmann \\
Department of Molecular Life Sciences,\\ Universit\"at Z\"urich\\
Z\"urich, Switzerland \\
\texttt{virginie.uhlmann@mls.uzh.ch}
}

\lmrlfinalcopy 
\begin{document}

\maketitle

\begin{abstract}
Although vision foundation models (VFMs) are increasingly reused for biomedical image analysis, it remains unclear whether the latent representations they provide are general enough to support effective transfer and reuse across heterogeneous microscopy image datasets. 
Here, we study this question for the problem of mitochondria segmentation in electron microscopy (EM) images, using two popular public EM datasets (Lucchi++ and VNC) and three recent representative VFMs (DINOv2, DINOv3, and OpenCLIP). 
We evaluate two practical model adaptation regimes: a frozen-backbone setting in which only a lightweight segmentation head is trained on top of the VFM, and parameter-efficient fine-tuning (PEFT) via Low-Rank Adaptation (LoRA) in which the VFM is fine-tuned in a targeted manner. 
Across all backbones, we observe that training on a single EM dataset yields good segmentation performance (quantified as foreground Intersection-over-Union), and that LoRA consistently improves in-domain performance. 
Training on the union of the two EM datasets does not degrade segmentation performance compared to single-dataset training, and LoRA provides further improvements across all backbones. 
Exploration of the latent representation space through various techniques (PCA, Fréchet Dinov2 distance, and linear probes) reveals a pronounced and persistent domain mismatch between the two considered EM datasets in spite of their visual similarity, which does not appear to prevent successful paired training. 
These results suggest that VFMs can support segmentation across heterogeneous EM datasets, even when their internal representations are not fully aligned. While DINO-based models generally outperform OpenCLIP, fully supervised end-to-end models can still achieve higher segmentation performance, highlighting the importance of selecting the appropriate architecture and training strategy for the target task and dataset. 
\end{abstract}

\section{Introduction}
Over the past decade, deep learning (DL) has become a central tool for quantitative microscopy image analysis, offering outstanding performance across tasks as diverse as image reconstruction, object segmentation, feature detection, and classification~\citep{moen2019deep, cunha2024machine}. 
At the same time, the ecosystem of publicly-available generalist models pre-trained on large datasets of diverse microscopy images has expanded rapidly, ranging from task-specific architectures trained for specific problems~\citep{schmidt2018cell, cellpose, weigert2018content} to large-scale vision and vision-language foundation models (FMs) intended to be general enough for broad reuse~\citep{zhao2025foundation, sam_m}. 
These developments have shifted the bottleneck for the adoption of DL approaches in microscopy image analysis: instead of designing new problem- and data-specific architectures, the practical hurdle now resides in determining \emph{when} and \emph{how} FMs can be safely used and, if required, adapted to new microscopy image data and biological questions.

A major challenge comes from the fact that microscopy data exhibit substantial distribution shift, even within the same imaging modality~\citep{stacke2020measuring}. 
Differences in sample preparation, acquisition protocols, resolution, contrast, and any other factor affecting the imaging itself can result in subtle dataset-specific signatures that hamper the performance generalization of DL models~\citep{abdullah2014sem}. 
As a result, it is hard to ensure that models are able to learn meaningful representations that remain stable across potentially big yet visually imperceptible domain shifts. 
This issue is also encountered in other, non-image data modalities, as illustrated by recent work questioning how ``foundational" FMs truly are for time-series forecasting, and demonstrating that generalization can collapse when the statistics of the original training and target domains are misaligned~\citep{karaouli2025foundational}. 
Motivated by this perspective, we here investigate an analogous question for vision foundation models (VFMs) used to analyse electron microscopy (EM) images, and explore whether current VFMs are foundational enough to support transfer across heterogeneous EM datasets or if they instead remain strongly domain-aware.
More specifically, we study transfer for the task of mitochondria segmentation in EM relying on two widely-used public datasets, Lucchi++~\citep{lucchi-dataset1, lucchi-dataset2} and VNC~\citep{droso-dataset}, and considering three recent representative VFMs, DINOv2~\citep{dinov2}, DINOv3~\citep{dinov3}, and OpenCLIP~\citep{openclip}. 
Rather than evaluating full end-to-end fine-tuning pipelines, we focus on two practical reuse settings that preserve most of the pre-trained weights: a frozen-backbone regime with a lightweight segmentation head, and parameter-efficient fine-tuning (PEFT) via low-rank adaptation (LoRA,~\cite{lora}).

We organize our analysis around three questions: (1) Can VFMs generalize beyond their pre-training distribution in EM image data, (2) can PEFT improve generalization across EM datasets, and (3) are VFMs with and without PEFT competitive with task-specific models and other end-to-end alternatives in practice. 
Across all backbones, we find that frozen-backbone adaptation performs well when training and testing on a single EM dataset, and that LoRA further improves in-domain performance. 
When training on the union of the two considered EM datasets, we observe that performance is generally maintained or improved compared to single-dataset training, with LoRA providing further improvements across all backbones.
Further representation-space diagnostics (principal component analysis (PCA), Fréchet distance computed in DINOv2 feature space~\citep{fdd}, and linear probes) indicate that Lucchi++ and VNC remain strongly separable even after PEFT, in spite of the good segmentation performance achieved under paired training.

\paragraph{Contributions.}
Our main contributions in this work are as follows: \textbf{(1)} we provide a controlled benchmark of three major VFMs on an EM segmentation task under frozen-backbone and LoRA-PEFT adaptation; \textbf{(2)} we provide empirical evidence that paired training across heterogeneous datasets from the same microscopy imaging modality can maintain comparable segmentation performance to single-dataset training; \textbf{(3)} we carry out domain-mismatch analyses to reveal persistent domain-specific representation separation across individual datasets despite successful joint training; \textbf{(4)} we report a performance comparison against representative non-VFM baselines from the EM segmentation literature.

\section{Related Work}\label{sec:rel-work}
Several works have recently started to explore whether FM paradigms can be meaningfully instantiated for specific biomedical imaging domains. 
For example, a recent multimodal model proposed for digital pathology has been explicitly positioning itself as an FM by integrating heterogeneous data modalities under a unified representation~\citep{ding2025multimodal}. 
In contrast, truly EM-native FMs remain relatively scarce. 
While a recent work frames itself as an FM approach for EM~\citep{yu2025emcf}, it only provides brief downstream evaluations. 
Furthermore, the reported performance of this model for mitochondria segmentation appears to be surprisingly low (mean IoU$<60\%$), suggesting that robust transfer for segmentation in EM remains an open challenge. 
In practice, the majority of the recent EM segmentation literature either still follows a paradigm in which task-specific models are trained end-to-end for a given dataset~\citep{franco2022stable}, or seeks to adapt general-purpose VFMs through fine-tuning and related strategies~\citep{sam_m, zhou2024mip}.

Recent efforts have investigated how pre-trained representations can be leveraged for EM segmentation, particularly for mitochondria. 
Two models have been proposed by~\cite{he2025unifying} specifically for mitochondria segmentation, namely EM-DINO, a DINO-style encoder pre-trained on EM data, and OmniEM, a unified dense-prediction model that builds on the EM-DINO embeddings. 
OmniEM reports improved mitochondria segmentation across diverse EM benchmarks, including the Lucchi++ dataset considered here. 
In parallel, DINOSim~\citep{dinosim} exploits DINOv2 patch embeddings and lightweight, non-parametric inference (\emph{e.g.}, similarity- and $k$NN-based labelling) to perform zero-shot semantic segmentation on EM datasets, including Lucchi++ and VNC.
The reported results highlight that fully zero-shot pipelines still offer a substantially poorer performance than supervised baselines, motivating the systematic study of when and why VFMs successfully transfer to EM. 
Beyond DINO-style encoders, other methods for self-supervised pre-training with GAN objectives on unlabelled EM data have been proposed and demonstrated consistent gains for downstream tasks, including semantic segmentation~\citep{kazimi2024self}. 
Finally, SAM-centric domain adaptation approaches such as SAMDA~\citep{wang2025samda} combine the Segment Anything Model (SAM)~\citep{sam2} with a U-Net~\citep{unet} or nnU-Net-style~\citep{nnu-net} expert component to improve transfer under limited supervision, further illustrating that domain shift remains a bottleneck even within a single microscopy imaging modality such as EM.

A particularly active line of work focuses on adapting SAM to biomedical image data~\citep{sam_m}, motivated by the observation that SAM’s off-the-shelf performance often degrades under microscopy-specific distributions. 
Recent studies have explored parameter-efficient adaptation of SAM, including LoRA-based tuning, for object segmentation in biomedical images~\citep{peftsam}.
Other approaches, such as mixtures of PEFT experts with gating mechanisms, have also been proposed to specialize SAM to different domains and tasks~\citep{sahay2025mopeft}.
In parallel, there is a growing trend towards the reuse of general-purpose VFMs trained on natural images as feature encoders for biomedical data. 
OpenCLIP embeddings have been used as a representative feature space to condition diffusion models for generating microscopy images~\citep{siemens2023synthetically}, and CLIP-style encoders have been adapted for downstream tasks such as multilabel classification in histopathology image data~\citep{bai2025histoclip}.
Similarly, DINO-family encoders are increasingly used in microscopy pipelines, as exemplified by DINOv2-based self-supervised models for 3D microscopy segmentation and tracking~\citep{lavaee2026spatialdino}, as well as by recent benchmarks evaluating DINOv3 as a transferable backbone across medical vision tasks~\citep{liu2025does}.
Given this growing body of work on SAM adaptation and domain-specific refinement, we chose not to centre our evaluation on SAM and instead focus on the comparatively less studied but widely deployed VFMs models DINOv2/DINOv3 and OpenCLIP to assess whether general-purpose pre-training on natural images can support robust EM segmentation and, crucially, whether lightweight adaptation suffices to bridge inter-dataset shifts within EM imaging.

\section{Methodology}\label{sec:meth}


\subsection{Datasets}
We use two EM datasets, VNC~\citep{droso-dataset} and Lucchi++~\citep{lucchi-dataset1, lucchi-dataset2}, selected for their wide adoption in the EM segmentation literature, which enables a direct comparison against a broad range of established methods for mitochondria segmentation in EM image data.
We provide in Supplementary Section~\ref{supp:dataset_overview} a detailed description of the content and format of these two datasets.

\subsection{Models}
\subsubsection{Vision Foundation Model Backbones}
We selected the $3$ widely used VFMs DINOv2~\citep{dinov2}, DINOv3~\citep{dinov3}, and OpenCLIP~\citep{openclip}, as they span complementary pre-training objectives (self-supervised vision pre-training and vision-language pre-training) and offer publicly available weights as well as robust, documented, and reproducible implementations. 
We provide an in-depth rationale for this selection in Supplementary Section~\ref{supp:backbones}. 
For all backbones, we primarily report results on the large version (L) of the underlying ViT model, although we observed that scaling it up or down does not have a major impact on results (Supplementary Section~\ref{supp:backbonescale}).

\subsubsection{Segmentation Head Architecture}\label{sec:seghead}
The considered VFM backbones tokenize an input image into non-overlapping patches and process a sequence of patch tokens, which we reshape into a 2D feature map of size $B \times D \times H' \times W'$, where $B$ is the batch size, $D$ is the backbone embedding dimension, and $(H', W')$ is the spatial patch grid, which depends on the (pre-processed) input resolution and on the model patch size $P \in \{14, 16\}$. 
The embedding dimension $D$ is not tuned independently, but is instead fixed by the chosen backbone variant (ViT-S/B/L/\dots) and corresponds to the token width of the vision encoder (pre-projection for OpenCLIP). 
For example, for the ViT-L variants used in our experiments, $D=1024$ for DINOv2/DINOv3, whereas OpenCLIP ViT-L/14 uses $D=768$ (with smaller variants typically using $D=384$ for ViT-S and $D=768$ for ViT-B).
We then attach a segmentation head composed of a lightweight convolutional decoder on top of these embeddings to obtain dense pixel-wise predictions.
This design keeps the segmentation head largely backbone-agnostic, as it only requires access to the patch-token grid and its channel dimension.

Importantly, we use the same segmentation head architecture and optimization protocol across all backbones and adaptation regimes.
The only backbone-specific change is the input-channel size to match the backbone embedding dimension $D$ and patch grid.
In the frozen-backbone setting, only the segmentation head parameters are optimized while the backbone is kept fixed.
Under PEFT, the segmentation head remains unchanged, and we additionally optimize the injected adapter parameters in the backbone. 
This ensures that performance differences across methods can be fully attributed to the backbone representations and adaptation strategy, rather than to changes in the model.

\subsection{Training and Adaptation Protocols}
We adopt a standardized training protocol across datasets and backbones to ensure that the observed differences in performance arise from the representation and adaptation regime rather than from optimization details. 
Unless stated otherwise, we optimize a Dice loss relying on the MONAI library implementation~\citep{monai} and train for up to $1000$ epochs with early stopping (patience$=20$), triggered by the validation loss. 
For each run, we select the checkpoint achieving the lowest validation loss and report test performance from that checkpoint. 
Validation splits are created by randomly holding out $10\%$ of the available training data. 
We use the AdamW~\citep{adamw} optimizer with learning rate $5\times10^{-5}$ and weight decay $10^{-4}$, and, following standard practice, apply no data augmentation as FMs are expected to have already learned robust features from their large-scale pre-training with extensive augmentation~\citep{dinov2}.
Due to the memory footprint of the backbones, we use a batch size of $2$ for all experiments. 
The number of iterations per epoch therefore depends on the size of the dataset, with $\lceil N_{\text{train}}/2 \rceil$ update steps per epoch.
We report results as the mean and standard deviation (std) averaged across multiple runs. 
We normalize the image inputs using ImageNet statistics (mean and std)~\citep{imagenet} for both training and validation, and do not apply any dataset-specific intensity normalization as expected for VFMs used in this setting.
All experiments were run on our institutional high-performance compute cluster using a single NVIDIA H100 GPU per run. 

\subsubsection{Frozen Backbone (Head-Only) Adaptation}\label{sec:headonly}
As a baseline, we adapt each VFM without updating its pre-trained backbone weights. 
Specifically, we keep the backbone frozen (including normalization layers), and only optimize the segmentation head parameters. 
Although this setting is not strictly ``zero-shot" (since it learns a task-specific decoder), it provides a low-capacity and comparable adaptation across backbones: the same segmentation head architecture, loss, optimizer, and training schedule are used, and the backbone representation is held fixed. 
This design allows us to isolate the contribution of the pre-trained representation quality to the final segmentation performance.

\subsubsection{PEFT with LoRA}\label{sec:peft}
To test whether limited backbone adaptation improves the robustness of performance under domain shift, we additionally employ LoRA~\citep{lora} as PEFT strategy. 
We provide an in-depth explanation of how LoRA is injected into the backbones in Supplementary Section~\ref{supp:lora_details}.
In this regime, the segmentation head is trained exactly as in the frozen-backbone baseline, while the trainable parameter set is augmented by the LoRA adapters.
While carrying out LoRA increases the total number of trainable parameters as compared to head-only adaptation, it remains relatively lightweight, especially considering the size of the VFM backbones (Supplementary Section~\ref{supp:parambudget}).

\subsubsection{Single Dataset vs.\ Paired Dataset Training Regimes}\label{sec:singlepaired}
We evaluate transfer under two training regimes: ``single dataset" training, where models are trained on Lucchi++ or VNC individually using their corresponding splits, and ``paired dataset" training, where models are trained on the union of both datasets to assess whether a single backbone can support generalized mitochondria segmentation in EM image data. 
For paired training, we intentionally use the natural (unbalanced) dataset mixture without reweighting or resampling to evaluate transfer under realistic data availability.
We always report performance on the test datasets, both in the single and paired dataset training scenarios.

A practical complication in the paired dataset training scenario is that the datasets have different native image dimensions and the backbones use different patch sizes. 
Instead of cropping the input data, we rescale each image to a backbone‑specific resolution by preserving its aspect ratio and setting the longest edge to a fixed target, then snapping both sides to a multiple of the patch size. 
This “longest‑edge with patch‑multiple” resizing strategy is applied only in the paired regime to avoid padding artifacts and yield inputs that are compatible with patch tokenization while preserving as much of the field of view as possible.

\subsection{Domain Mismatch Analysis}
To investigate the model's internal representations of Lucchi++ and VNC, we performed two lightweight diagnostics in the latent space of the backbone VFM. 
For each image, we extract a fixed-dimensional embedding from the VFM (either obtained with frozen or PEFT-adapted weights) by pooling the final-layer patch tokens into a single feature vector. 
In our implementation, we used a simple global pooling operation on the patch token grid, but observed that other standard choices (\emph{e.g.}, using the CLS token instead) lead to the same qualitative conclusions.

\paragraph{PCA visualization and Fréchet DINOv2 distance.}
We first visualize the joint embedding space by applying PCA to the pooled features and projecting both datasets onto the first two leading principal components. 
To quantify distributional mismatch, we compute a Fréchet distance between the two feature distributions in the same spirit as the Fréchet Inception Distance~\citep{fid}, but using DINOv2 features instead of Inception features, hence referred to as Fréchet DINOv2 Distance (FD-DINOv2,~\cite{fdd}, Supplementary Section~\ref{supp:metrics}). 
We then report FD-DINOv2 for the frozen and LoRA-adapted backbones to assess whether PEFT reduces the representation gap.

\paragraph{Linear separability of domains.}
As a complementary readout, we also train a logistic regression (LR) classifier to predict the dataset label (Lucchi++ vs.\ VNC) from the extracted embeddings, using the same train/validation split strategy as in the segmentation experiments. 
We then compare the performance of this linear probe for features extracted with the frozen and LoRA-adapted backbones to determine whether PEFT meaningfully decreases domain discriminability.

\subsection{Performance Evaluation Metrics}
We computed performance metrics over the full evaluation split in a single pass using \emph{dataset-level} pixel totals 
rather than averaging per-image scores to avoid the high variance that could arise when mitochondria occupy very different fractions of the overall image across slices. 
We evaluate segmentation performance using the foreground IoU (IoU$_\mathrm{fg}$, Supplementary Section~\ref{supp:metrics}), where all non-zero labels are collapsed into a single foreground class. 
The foreground IoU is a standard metric in the EM segmentation literature to avoid biasing the metric towards the background class, which is easier to predict in images where the foreground is sparse~\citep{iou_fg}. 
It also enables direct comparison to prior work on EM segmentation~\citep{dinosim}. 

\section{Results}\label{sec:results}

\subsection{Generalization of VFMs to EM image data}\label{sec:res1}
To probe the extent to which general-purpose VFMs transfer to EM, we first benchmarked our $3$ backbones under a frozen-backbone, segmentation head-only adaptation regime on Lucchi++ and VNC as described in~\ref{sec:headonly}. 
In Table~\ref{tab:frozen_results}, we report the IoU$_\mathrm{fg}$ for the three training settings outlined in~\ref{sec:singlepaired} (Lucchi++ only, VNC only, and paired training).

Across single-dataset training, all models achieve a relatively strong IoU$_\mathrm{fg}$ on both Lucchi++ and VNC considering the nature of the frozen-backbone approach, indicating that pre-trained representations can support effective adaptation to an individual EM dataset.
Despite Lucchi++ being a FIB-SEM and VNC being an ssTEM dataset, paired training maintains comparable performance to single-dataset training across all backbones.
To test whether the results in the paired training scenario were affected by dataset imbalance between Lucchi++ and VNC, we reran paired training with balanced sampling (Supplementary Section~\ref{supp:unbalanced}). 
Our results indicate that balanced sampling does not improve performance, with comparable macro IoU$_\mathrm{fg}$ values for balanced and unbalanced training.
\begin{table}[t]
\centering
\small
\setlength{\tabcolsep}{8pt}
\renewcommand{\arraystretch}{1.15}
\begin{tabular}{lccc}
\toprule
\textbf{Backbone} & \textbf{Lucchi++} & \textbf{VNC} & \textbf{Paired} \\
\midrule
DINOv2-L/14 & $\mathbf{0.717 \pm 0.006}$ & $0.723 \pm 0.002$ & \changed{$0.717 \pm 0.003$} \\
DINOv3-L/16 & $\mathbf{0.717 \pm 0.000}$ & $\mathbf{0.833 \pm 0.000}$ & $\changed{\mathbf{0.793 \pm 0.002}}$ \\
OpenCLIP ViT-L/14 & $0.568 \pm 0.003$  & $0.573 \pm 0.011$ & \changed{$0.582 \pm 0.019$} \\
\bottomrule
\end{tabular}
\caption{\textbf{Frozen-backbone (head-only) segmentation performance.} 
Foreground IoU (IoU$_\mathrm{fg}$) on Lucchi++ and VNC under single-dataset and paired training, reported as the mean$\pm$std over $5$ runs. 
In the paired setting, the reported IoU$_\mathrm{fg}$ is a macro-average across both datasets.}
\label{tab:frozen_results}
\end{table}

\paragraph{Domain mismatch within the same modality.}
To investigate the model's internal representation of Lucchi++ and VNC, we analyzed the backbone feature space for both datasets with DINOv2-L/14. 
In Figure~\ref{fig:pca_mismatch_a}, we show a PCA projection of the image-level embeddings extracted from the frozen backbone. 
The two datasets occupy disjoint regions of the representation space, indicating that the pre-trained features encode strong dataset-specific signatures rather than a unified EM images manifold. 
This qualitative separation is corroborated quantitatively by a large Fréchet distance ($79.32$) between Gaussian approximations of the two feature distributions (FD-DINOv2), and by a near-perfect linear separability of the dataset label using an LR probe (accuracy and ROC-AUC of $1.0$).
Further analysis of the unlabelled stack (stack $2$) of VNC yields similar results, indicating that these observations cannot be attributed to small-sample effects (Supplementary Section~\ref{supp:vncstack2}).
Together, these diagnostics indicate substantial domain mismatch between Lucchi++ and VNC at the representation level, which does not seem to prevent the model from maintaining comparable segmentation performance when trained on the two datasets simultaneously.

\subsection{Improvement of generalization across EM datasets through PEFT}\label{sec:res2}
We next tested whether PEFT can further improve the performance of VFMs under paired training. 
Using the same experimental protocol as in the previous section, we enable LoRA adapters while keeping the segmentation head unchanged.
In Table~\ref{tab:lora_results}, we report the IoU$_\mathrm{fg}$ for single-dataset and paired training.

\begin{table}[t]
\centering
\small
\setlength{\tabcolsep}{8pt}
\renewcommand{\arraystretch}{1.15}
\begin{tabular}{lccc}
\toprule
\textbf{Backbone} & \textbf{Lucchi++} & \textbf{VNC} & \textbf{Paired} \\
\midrule
DINOv2-L/14        & $0.802 \pm 0.007$ & $0.818 \pm 0.003$ & \changed{$0.826 \pm 0.007$} \\
DINOv3-L/16        & $\mathbf{0.838 \pm 0.002}$ & $\mathbf{0.875 \pm 0.001}$ & $\changed{\mathbf{0.847 \pm 0.003}}$ \\
OpenCLIP ViT-L/14  & $0.624 \pm 0.007$ & $0.679 \pm 0.017$ & \changed{$0.728 \pm 0.006$} \\
\bottomrule
\end{tabular}
\caption{\textbf{LoRA-PEFT segmentation performance.} Foreground IoU (IoU$_\mathrm{fg}$) on Lucchi++ and VNC under single-dataset and paired training with LoRA adapters in the backbone, reported as the mean$\pm$std over $5$ runs. 
In the paired setting, the reported IoU$_\mathrm{fg}$ is a macro-average across both datasets.}
\label{tab:lora_results}
\end{table}

\begin{figure}[t]
    \centering
    \begin{subfigure}[t]{0.45\linewidth}
        \centering
        \includegraphics[width=\linewidth]{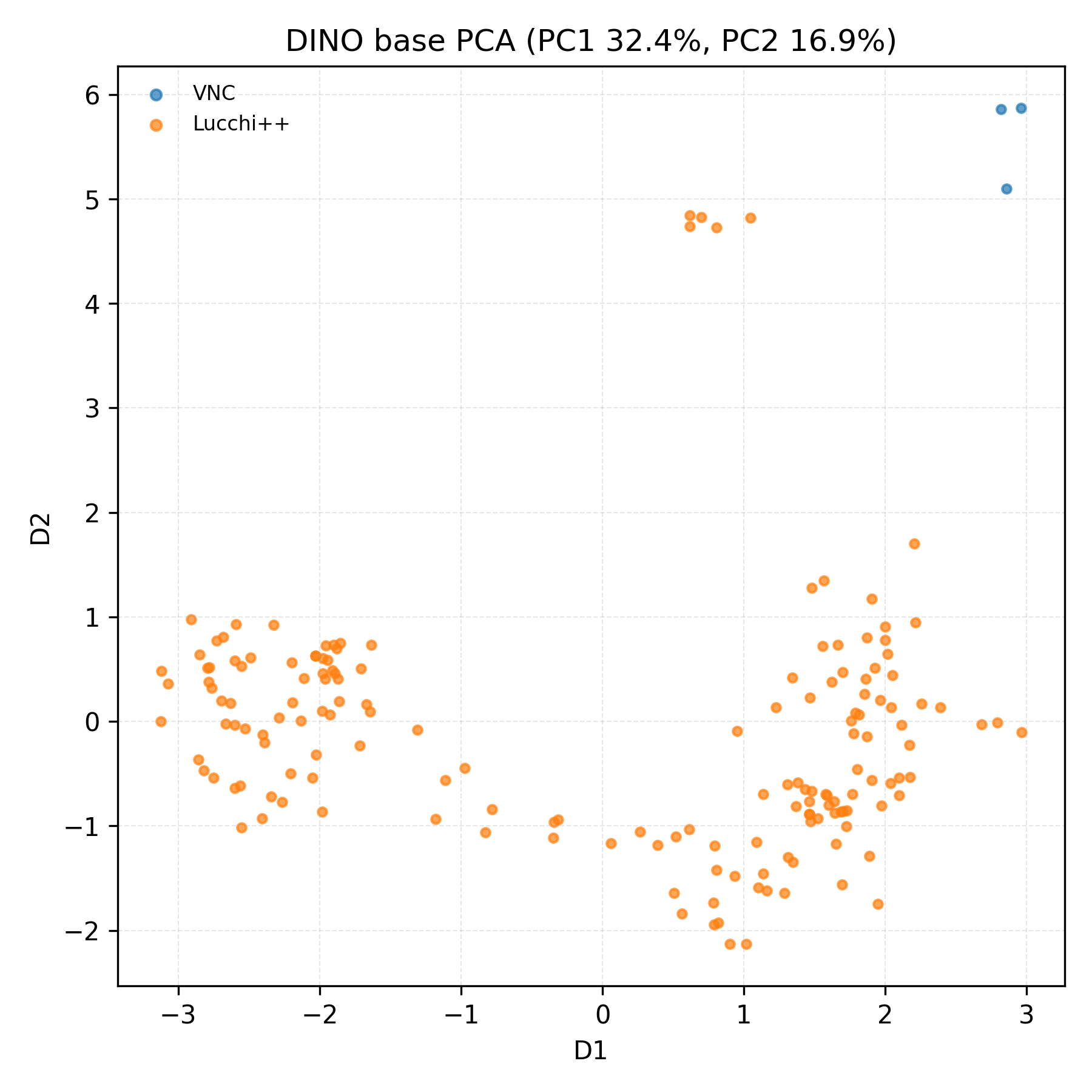}
        \caption{Frozen backbone (head-only training)}
        \label{fig:pca_mismatch_a}
    \end{subfigure}
    \hfill
    \begin{subfigure}[t]{0.45\linewidth}
        \centering
        \includegraphics[width=\linewidth]{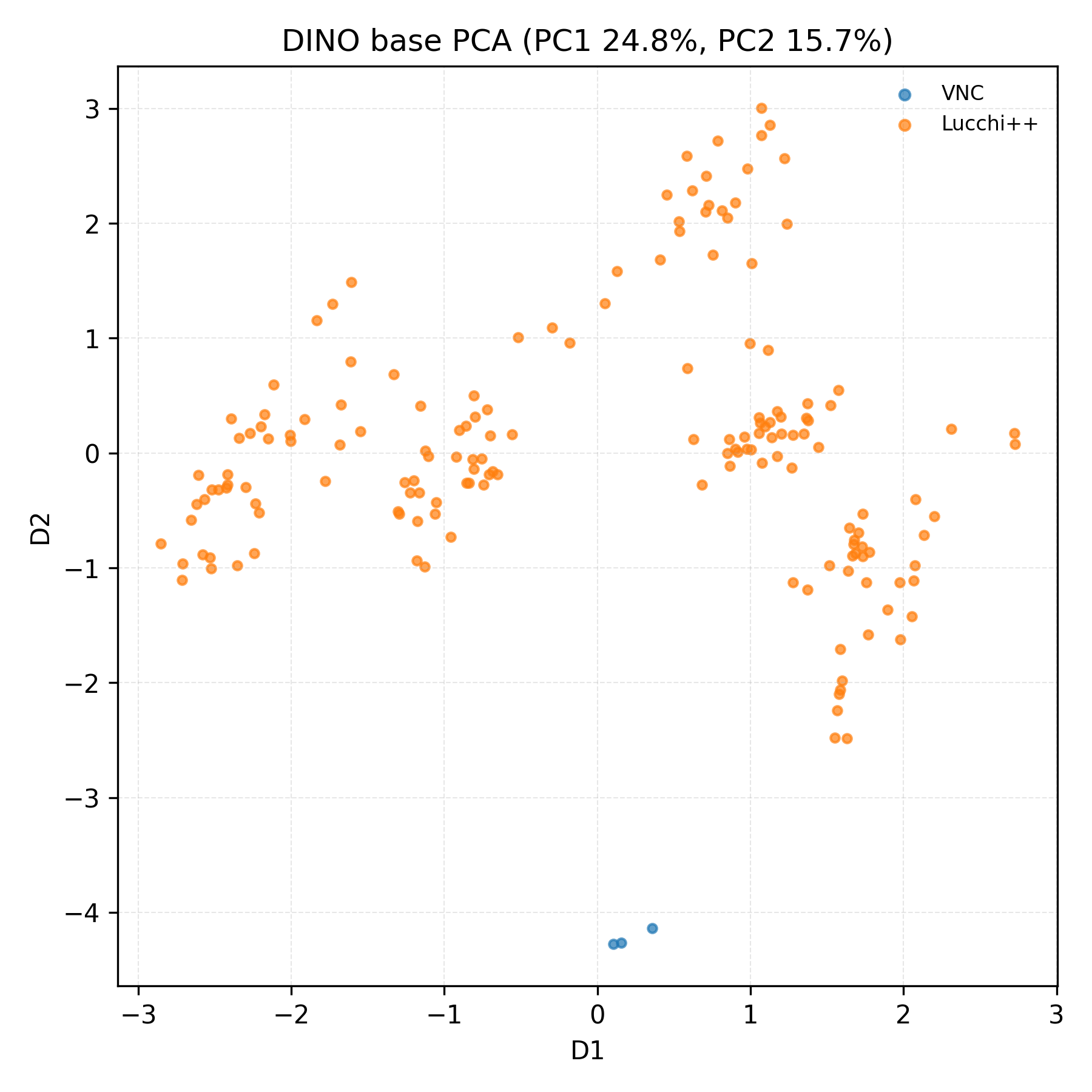}
        \caption{LoRA-PEFT backbone}
        \label{fig:pca_mismatch_b}
    \end{subfigure}

    \caption{\textbf{Domain mismatch between Lucchi++ and VNC in DINOv2 feature space.}
(a) PCA projection of image-level embeddings from the frozen DINOv2-L/14 backbone exhibits no overlap between the two datasets. 
(b) With LoRA enabled, the feature-space gap is minimally reduced, and the two domains remain clearly separable. 
We also report the FD-DINOv2 (\ref{eq:fddino}) and LR performance for predicting the dataset label from the embedding vectors (frozen: $79.32$; LoRA: $45.02$; probe accuracy/AUROC $=1.0$ in both).}
    \label{fig:pca_mismatch_both}
\end{figure}

Across single-dataset training, LoRA consistently improves performance for all backbones by approximately $5-12$ points relative to the frozen setting, indicating that limited backbone adaptation is beneficial when the training and test data come from a single EM source. 
Similarly, paired training also benefits from LoRA, with noticeable improvements in macro  IoU$_\mathrm{fg}$ across all backbones. 
These results suggest that PEFT can improve segmentation performance both when training on individual EM datasets and when training jointly on heterogeneous EM datasets.

\paragraph{Domain mismatch is reduced but persists after PEFT.}
To assess whether LoRA indeed aligns the two datasets in representation space, we repeated the domain mismatch diagnostics from~\ref{sec:res1} using embeddings extracted from the LoRA-adapted backbone. 
In Figure~\ref{fig:pca_mismatch_b}, we observe that Lucchi++ and VNC are brought slightly closer from one another in the PCA projection as compared to the frozen backbone scenario, but remain clearly separable. 
Quantitatively, although the Fréchet distance computed in the DINOv2 feature space decreases from $79.32$ (frozen) to $45.02$ (LoRA), an LR probe is still able to predict the dataset label near-perfectly (accuracy/AUROC $=1.0$).
Further analysis of the unlabelled stack of VNC yields similar results, indicating that these observations cannot be attributed to small-sample effects (Supplementary Section~\ref{supp:vncstack2}).
Together, these results indicate that LoRA reduces but does not eliminate the representation gap between Lucchi++ and VNC, while still improving segmentation performance under paired training.

\subsection{Performance of VFMs with and without PEFT against state-of-the-art alternatives}
\begin{figure}[t]
    \centering
    \includegraphics[width=0.9\linewidth]{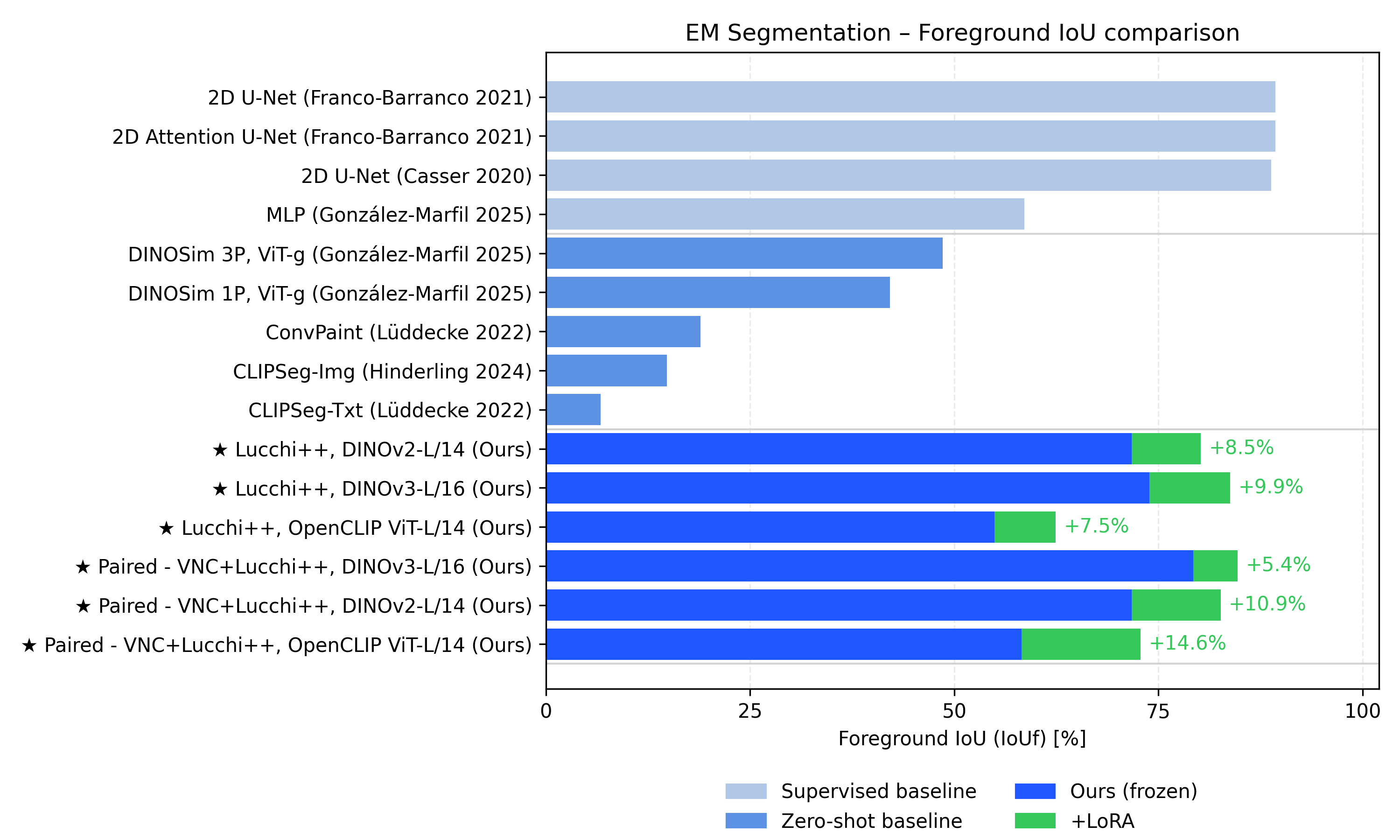}
    \caption{\textbf{Comparison to prior EM mitochondria segmentation approaches.} Foreground IoU (IoU$_\mathrm{fg}$) for supervised baselines (top, light blue), zero-shot and prompt-based methods (middle, blue), and our VFM-based models with frozen (bottom, dark blue) and LoRA-adapted backbones (bottom, green). 
    Supervised baselines were trained for mitochondria segmentation on Lucchi++ dataset.
    Our experiments use the large backbone variants (DINOv2-L/14, DINOv3-L/16, and OpenCLIP ViT-L/14) reported in the Results section. While fully supervised baselines generally achieve higher performance, VFMs remain competitive under both single-dataset and paired training, with further improvements obtained with LoRA.}
    \label{fig:sota_comp}
\end{figure}

Finally, we compare the quality of the VFM-based segmentation against representative state-of-the-art alternatives proposed in previous works on mitochondria segmentation on EM. 
In Figure~\ref{fig:sota_comp}, we report the IoU$_\mathrm{fg}$ for supervised end-to-end baselines (\emph{e.g.}, U-Net variants), prior prompt-based and zero-shot approaches (\emph{e.g.}, CLIPSeg, ConvPaint), and for our VFM-based models with and without LoRA. 
To facilitate comparison with results reported in the literature, we have only evaluated our single-dataset work on Lucchi++ and reserved the VNC dataset for the paired experiments. 

As expected, fully supervised end-to-end baselines remain the strongest performers as they are optimized directly for their target dataset and task. 
In contrast, VFM-based models, even when enhanced with LoRA, do not manage to consistently match the best U-Net variants. 
Importantly, however, this performance gap comes with different computational and practical requirements: while end-to-end training requires substantially more task-specific optimization and compute, frozen-backbone adaptation and PEFT offer a general way to only update a small fraction of parameters while reusing the pre-trained representation. 
From this perspective, VFMs can offer an attractive option when compute or annotation budgets are constrained, or when quick adaptation to a new dataset is desired.

The paired setting, however, highlights the potential of VFMs to support training across heterogeneous EM datasets. 
While these models can transfer reasonably well to individual microscopy image datasets despite being pre-trained predominantly on natural images, they can also support a single model trained on the union of different EM imaging datasets. 
In our experiments, we observe that performance under paired training is generally maintained or improved compared to the single-dataset setting, and that LoRA provides further improvement across all backbones. 
This suggests that, besides the mismatch between natural images and EM data and the intra-modality differences between the EM datasets, VFMs can provide useful representations for segmentation across heterogeneous microscopy imaging datasets.
Moreover, the persistent separation between datasets in representation space does neither prevent successful paired training nor improvements through lightweight adaptation.

\section{Conclusion}\label{sec:conclusion}
Across three widely used VFMs (DINOv2, DINOv3 and OpenCLIP), we found that pre-trained representations can be effectively reused for mitochondria segmentation on EM image data when training and testing within a single dataset.
While a simple decoder added on a frozen backbone already yields strong performance, a refinement of the backbone with LoRA further improves in-domain performance. 
When training a single ``general EM'' model on the union of two EM datasets, we observe that segmentation performance is generally maintained or improved compared to single-dataset training, with PEFT providing further improvements across all considered backbones. 
Investigations of the latent representations of the models considered relying on PCA projections, calculation of the Fréchet distance in DINOv2 feature space, and linear probing, indicate that the EM datasets remain strongly separable even after PEFT in spite of their comparable segmentation performance under paired training. 
This suggests that reducing dataset separability is not necessary to improve downstream segmentation performance.
Overall, our results suggest that current VFMs can support segmentation across heterogeneous EM datasets, even when the datasets remain distinguishable in the representation space.
They also highlight that adaptation strategies such as LoRA can further improve performance without requiring full fine-tuning of the backbone.
However, the benefits of VFMs depend on the choice of backbone, adaptation strategy, and characteristics of the target data and task.
In our experiments, DINO-based models generally outperform OpenCLIP, while fully supervised end-to-end approaches remain competitive and can achieve higher segmentation performance.
These observations support the potential of VFMs to serve as general models for microscopy image analysis



\subsubsection*{Acknowledgments}
The authors were supported by the University of Zurich.

\bibliography{iclr2025_conference}
\bibliographystyle{iclr2025_conference}

\appendix
\section{Supplementary Material}
\subsection{Dataset overview}\label{supp:dataset_overview}
\paragraph{VNC (Drosophila ssTEM)~\citep{droso-dataset}.}
The VNC dataset is commonly used to study mitochondria segmentation. 
It consists of two sequential serial-section transmission electron microscopy (ssTEM) image stacks of the \emph{Drosophila melanogaster} third-instar larva ventral nerve cord (VNC), each composed of $1024\times1024\times20$ voxels.
The first stack provides multiple object labels, including binary mitochondria segmentation masks (foreground mitochondria vs.\ background), while the second one contains raw images only.
Following prior work using this dataset~\citep{dinosim}, we used the annotated stack as our supervised benchmark and split its $20$ slices into $17$ training and $3$ test images. 
The second (unlabelled) stack was reserved for unsupervised analyses, as it does not include any ground-truth annotations.

\paragraph{Lucchi++ (Mouse Hippocampus FIB-SEM)~\citep{lucchi-dataset1, lucchi-dataset2}.}
The Lucchi++ dataset is a focused ion beam scanning electron microscopy (FIB-SEM) image volume of the mouse hippocampus (CA1) acquired at an isotropic $5$nm resolution, and is commonly used for mitochondria segmentation benchmarks. 
In contrast to VNC, the Lucchi++ dataset comes with a predefined train/test split and corresponding annotations. 
Specifically, the dataset includes two sequential stacks for training and testing, with $165$ images per split (\emph{i.e.}, $165$ annotated training images and $165$ annotated test images), yielding four sequential stacks in total. 
Each slice is composed of $1024\times768$ pixels and is accompanied by a binary mitochondria segmentation mask (foreground mitochondria vs.\ background), which we use as sole supervision signal in our experiments.
We adopt this standard split throughout our experiments with this dataset.

\begin{figure}
    \centering
    \includegraphics[width=0.9\linewidth]{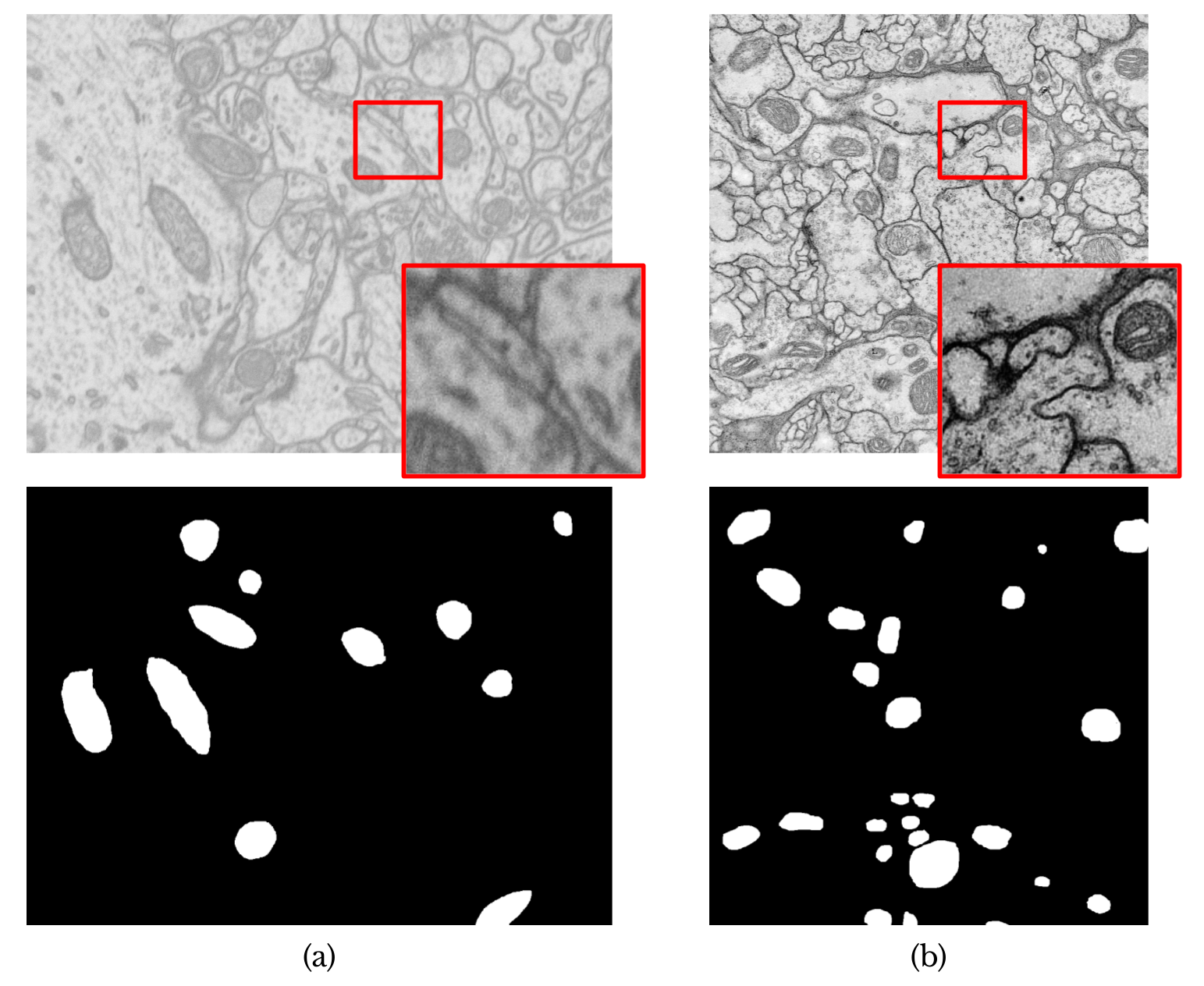}
    \caption{\textbf{Illustration of the two EM datasets used in this work.}
    Representative slices from (a) Lucchi++ (mouse hippocampus, FIB-SEM) and (b) VNC (Drosophila ventral nerve cord, ssTEM), shown with their corresponding binary mitochondria ground-truth masks.
    The red box and inset highlight a local region to emphasize differences in texture/contrast and membrane appearance across datasets, despite both being EM images.}
    \label{fig:dataset_examples}
\end{figure}

In Figure~\ref{fig:dataset_examples}, we illustrate representative images of the two datasets.

\subsection{Rationale for the choice of Vision Foundation Model Backbones}\label{supp:backbones}
We here motivate the choice of backbones for transfer to our microscopy image data analysis application of focus.

First, we selected DINOv2~\citep{dinov2}, which appeared as a natural starting point as recent work reported promising performance on EM segmentation leveraging DINOv2 features, albeit under a different adaptation strategy than ours~\citep{dinosim}. 

Second, considering the recent release of DINOv3~\citep{dinov3} and its emerging adoption in microscopy image analysis settings through parameter-efficient adaptation on other modalities~\citep{cytodino, peftdinov3-mitotic}, we tested whether this newer model improves the robustness and transfer of the data representation as compared to DINOv2 under the same controlled protocol.
Finally, we also included OpenCLIP~\citep{openclip} to probe a qualitatively different pre-training signal. 

OpenCLIP has not been broadly studied for EM segmentation, and only limited benchmarking evidence in microscopy data is available with this model to date~\citep{micro-bench}.
It is a vision-language architecture, although we do not consider the language part in this work.
As such, it offers an opportunity to evaluate whether a backbone with such a different architecture provides any advantage (or failure mode) under microscopy data domain shift.

All three backbones are pre-trained on a large corpus of natural images (\emph{i.e.}, not microscopy data), but with different objectives and data sources: DINOv2 is trained in a self-supervised manner on the curated LVD-142M web dataset, DINOv3 is trained on a curated subset derived from a multi-billion pool of public web images, and the OpenCLIP variant we use is trained contrastively on the LAION-2B subset of LAION-5B.
Our working hypothesis is that the scale and diversity of these pre-training paradigms yield broadly reusable visual primitives and robust representations that can transfer to microscopy images with limited supervision.
Our experiments aim to explicitly test this assumption and to assess the extent to which parameter-efficient adaptation facilitates the transfer to EM imaging data.

All backbones are ViT-based and thus operate on a fixed patch size, which determines the token grid and constrains input handling. 
DINOv2 is released as ViT-$\cdot$/14, and we therefore standardize our preprocessing to be compatible with a $14\times14$ patchification, as described in Section~\ref{sec:seghead}. 
DINOv3 uses ViT-$\cdot$/16 and its implementation is more permissive to varying input sizes (\emph{e.g.}, through positional-embedding handling), reducing the need for strict input divisibility. 
OpenCLIP is available in both /14 and /16 variants.
Since our primary OpenCLIP backbone is ViT-L/14 and to avoid confounding tokenization geometry with model choice, we consistently use the /14 OpenCLIP variants. 

\subsection{Effect of backbone scale on segmentation performance}\label{supp:backbonescale}
In Tables~\ref{tab:supp_dinov2_variants},~\ref{tab:supp_dinov3_variants}, and~\ref{tab:supp_openclip_variants}, we report the IoU$_\mathrm{fg}$ for multiple backbone sizes (ViT-S: small, ViT-B: big, ViT-L: large, and ViT-G: giant) under the two adaptation regimes considered (frozen-backbone training and LoRA-PEFT). 
Across all three analyzed VFMs, scaling the backbone generally yields incremental gains in the single-dataset setting, and LoRA consistently improves performance compared to the frozen baseline for a fixed backbone size. 
Under paired training, performance is generally maintained or improved compared to single-dataset training across backbone sizes, with LoRA providing further improvements in most configurations. These results suggest that increasing backbone size is not necessary to maintain segmentation performance when combining EM datasets, although larger models still generally allow better absolute performances.

\begin{table}[ht]
\centering
\resizebox{\textwidth}{!}{
\begin{tabular}{lcccccc}
\toprule
\textbf{DINOv2} &
\multicolumn{2}{c}{\textbf{Lucchi++ (IoU$\_\mathrm{fg}$)}} &
\multicolumn{2}{c}{\textbf{VNC (IoU$\_\mathrm{fg}$)}} &
\multicolumn{2}{c}{\textbf{Paired (IoU$\_\mathrm{fg}$)}} \\
\cmidrule(lr){2-3}\cmidrule(lr){4-5}\cmidrule(lr){6-7}
& \textbf{Frozen} & \textbf{LoRA} & \textbf{Frozen} & \textbf{LoRA} & \textbf{Frozen} & \textbf{LoRA} \\
\midrule
ViT-S/14 & $0.606 \pm 0.008$ & $0.731 \pm 0.007$ & $0.769 \pm 0.004$ & $0.825 \pm 0.001$ & \changed{$\mathbf{0.699 \pm 0.008}$} & \changed{$\mathbf{0.803 \pm 0.003}$} \\
ViT-B/14 & $0.716 \pm 0.009$ & $0.774 \pm 0.005$ & $0.757 \pm 0.006$ & $0.857 \pm 0.004$ & \changed{$0.749 \pm 0.005$} & \changed{$0.821 \pm 0.009$} \\
ViT-L/14 & $0.717 \pm 0.006$ & $0.802 \pm 0.007$ & $0.723 \pm 0.002$ & $0.818 \pm 0.002$ & \changed{$0.717 \pm 0.003$} & \changed{$\mathbf{0.826 \pm 0.007}$} \\
ViT-G/14 & $\mathbf{0.783 \pm 0.002}$ & $\mathbf{0.852 \pm 0.003}$ & $\mathbf{0.810 \pm 0.003}$ & $\mathbf{0.860 \pm 0.003}$ & \changed{$0.784 \pm 0.003$} & \changed{$0.851 \pm 0.002$} \\
\bottomrule
\end{tabular}}
\caption{\textbf{Segmentation performance of DINOv2 variants.} Foreground IoU (IoU$\_\mathrm{fg}$) in the frozen-backbone (head-only) and LoRA-PEFT setting reported as the mean$\pm$std over repeated runs.}
\label{tab:supp_dinov2_variants}
\end{table}

\begin{table}[ht]
\centering
\resizebox{\textwidth}{!}{
\begin{tabular}{lcccccc}
\toprule
\textbf{DINOv3} &
\multicolumn{2}{c}{\textbf{Lucchi++ (IoU$_\mathrm{fg}$)}} &
\multicolumn{2}{c}{\textbf{VNC (IoU$_\mathrm{fg}$)}} &
\multicolumn{2}{c}{\textbf{Paired (IoU$_\mathrm{fg}$)}} \\
\cmidrule(lr){2-3}\cmidrule(lr){4-5}\cmidrule(lr){6-7}
& \textbf{Frozen} & \textbf{LoRA} & \textbf{Frozen} & \textbf{LoRA} & \textbf{Frozen} & \textbf{LoRA} \\
\midrule
ViT-S/16 & $0.598 \pm 0.000$ & $0.722 \pm 0.009$ & $0.772 \pm 0.000$ & $0.831 \pm 0.002$ & \changed{$0.706 \pm 0.006$} & \changed{$\mathbf{0.814 \pm 0.001}$} \\
ViT-B/16 & $0.646 \pm 0.000$ & $0.775 \pm 0.004$ & $0.772 \pm 0.000$ & $0.828 \pm 0.002$ & \changed{$0.711 \pm 0.012$} & \changed{$0.817 \pm 0.007$} \\
ViT-L/16 & $\mathbf{0.717 \pm 0.000}$ & $\mathbf{0.838 \pm 0.002}$ & $\mathbf{0.833 \pm 0.000}$ & $\mathbf{0.875 \pm 0.001}$ & \changed{$\mathbf{0.793 \pm 0.002}$} & \changed{$0.847 \pm 0.003$} \\
\bottomrule
\end{tabular}}
\caption{\textbf{Segmentation performance of DINOv3 variants.} Foreground IoU (IoU$_\mathrm{fg}$) in the frozen-backbone (head-only) and LoRA-PEFT setting reported as the mean$\pm$std over $5$ repeated runs.}
\label{tab:supp_dinov3_variants}
\end{table}

\begin{table}[ht]
\centering
\resizebox{\textwidth}{!}{
\begin{tabular}{lcccccc}
\toprule
\textbf{OpenCLIP} &
\multicolumn{2}{c}{\textbf{Lucchi++ (IoU$_\mathrm{fg}$)}} &
\multicolumn{2}{c}{\textbf{VNC (IoU$_\mathrm{fg}$)}} &
\multicolumn{2}{c}{\textbf{Paired (IoU$_\mathrm{fg}$)}} \\
\cmidrule(lr){2-3}\cmidrule(lr){4-5}\cmidrule(lr){6-7}
& \textbf{Frozen} & \textbf{LoRA} & \textbf{Frozen} & \textbf{LoRA} & \textbf{Frozen} & \textbf{LoRA} \\
\midrule
ViT-L/14 & $\mathbf{0.568 \pm 0.003}$ & $\mathbf{0.624 \pm 0.007}$ & $0.573 \pm 0.011$ & $0.679 \pm 0.017$ & \changed{$0.582 \pm 0.019$} & \changed{$0.728 \pm 0.006$} \\
ViT-H/14 & $0.556 \pm 0.003$ & $0.610 \pm 0.011$ & $\mathbf{0.578 \pm 0.008}$ & $\mathbf{0.687 \pm 0.012}$ & \changed{$\mathbf{0.579 \pm 0.027}$} & \changed{$\mathbf{0.718 \pm 0.011}$} \\
\bottomrule
\end{tabular}}
\caption{\textbf{Segmentation performance of OpenCLIP variants.} Foreground IoU (IoU$_\mathrm{fg}$) in the frozen-backbone (head-only) and LoRA-PEFT setting reported as the mean$\pm$std over repeated runs.}
\label{tab:supp_openclip_variants}
\end{table}

\subsection{LoRA-PEFT configuration and trainable parameters}\label{supp:lora_details}

To make our PEFT setting fully explicit, we here detail the specific backbone parameters we adapt and how. 
Across all backbones and experimental regimes, the segmentation head architecture and its training procedure are kept fixed.
The only additional trainable parameters introduced under PEFT are the LoRA adapter weights injected inside the backbone. 
Concretely, we apply LoRA to the ViT attention projections only (query/key/value and output projection), while keeping all original backbone weights frozen.
We provide a summary of the details of the PEFT configuration for each backbone in Table~\ref{tab:peft_config}.

\paragraph{Trainable vs.\ frozen parameters under PEFT.}
Under PEFT, the set of trainable parameters consists of the segmentation head parameters and the low-rank LoRA matrices $\mathbf{A}$ and $\mathbf{B}$ attached to each targeted linear projection. 
All original backbone weights remain frozen after LoRA injection.
Formally, for a targeted linear map with weight $\mathbf{W}$, LoRA replaces it with
$\mathbf{W}'=\mathbf{W} + \frac{\alpha}{r}\mathbf{B}\mathbf{A}$,
where $r$ is the LoRA rank, $\alpha$ is the scaling factor, and $\mathbf{A},\mathbf{B}$ are the two LoRA matrices being optimized.

\paragraph{LoRA hyperparameters and patch size.}
In all runs, we used $r=16$ based on small preliminary sweeps and following common practice in the PEFT literature.
Increasing $r$ is typically expected to improve performance but also increases the number of trainable parameters linearly, bringing the method closer to full fine-tuning and reducing the parameter-efficiency relevance~\citep{lora, qlora}. 
Similarly, we set the scaling $\alpha=32$ following standard LoRA conventions to stabilize the effective update magnitude across ranks~\citep{agiza2024mtlora, huggingface}.

The backbone patch sizes are fixed by the pre-trained model variants: DINOv2 uses patch size $14$, DINOv3 uses patch size $16$, and the OpenCLIP variants used in this work use patch size $14$.

\begin{table}[t]
\centering
\resizebox{\textwidth}{!}{
\begin{tabular}{lccccccc}
\toprule
\textbf{Backbone} & \textbf{Patch size} & \textbf{LoRA targets} & $r$ & $\alpha$  & \textbf{Trainable parameters} & \textbf{Frozen parameters} \\
\midrule
DINOv2      & $14$ & attn qkv + attn proj & $16$ & $32$  & head + LoRA $\mathbf{A},\mathbf{B}$ & backbone \\
DINOv3      & $16$ & attn qkv + attn proj & $16$ & $32$ & head + LoRA $\mathbf{A},\mathbf{B}$ & backbone \\
OpenCLIP        & $14$ & attn qkv + attn proj & $16$ & $32$  & head + LoRA $\mathbf{A},\mathbf{B}$ & backbone \\
\bottomrule
\end{tabular}}
\caption{\textbf{PEFT configurations by backbone.} 
We apply LoRA to attention projections only (qkv and output projection). 
Under PEFT, we only optimize the segmentation head and LoRA adapter matrices, while all the original backbone weights remain frozen.}
\label{tab:peft_config}
\end{table}

\paragraph{Parameter count reporting.}
For transparency, each run writes a \texttt{lora\_targets.json} report containing the resolved LoRA targets and the corresponding trainable parameter counts (total trainable, LoRA-only, and head-only). 
We used these reports to verify that only the intended modules were optimized.

\subsection{Trainable parameter budget under lightweight adaptation}\label{supp:parambudget}
To contextualize the accuracy-efficiency trade-off of lightweight adaptation, we report in Table~\ref{tab:supp_trainable_params_all} the number of trainable parameters for head-only training versus LoRA-PEFT (LoRA adapters plus segmentation head) for several backbones. 
As expected, head-only adaptation keeps the trainable parameter budget nearly constant across backbone sizes, since only the decoder head is optimized. 
In contrast, enabling LoRA increases the trainable parameter count with model scale. 
These numbers highlight that PEFT updates only a small fraction of the backbone while still yielding consistent in-domain gains, as seen in Tables~\ref{tab:supp_dinov2_variants} and~\ref{tab:supp_dinov3_variants}.

\begin{table}[ht]
\centering
\resizebox{\textwidth}{!}{
\begin{tabular}{lccc}
\toprule
\textbf{Backbone} & \textbf{Variant} &
\textbf{Trainable parameters (head-only)} &
\textbf{Trainable parameters (LoRA+head)} \\
\midrule
DINOv2  & ViT-S/14  & $3.829$M (14.8\%) & $4.271$M (16.5\%) \\
DINOv2  & ViT-B/14  & $4.025$M (4.4\%)  & $4.910$M (5.4\%) \\
DINOv2  & ViT-L/14  & $4.157$M (1.3\%)  & $6.516$M (2.1\%) \\
DINOv2  & ViT-G/14  & $4.419$M (0.4\%)  & $10.317$M (0.9\%) \\
\midrule
DINOv3  & ViT-S/16  & $3.829$M (15.1\%) & $4.271$M (16.8\%) \\
DINOv3  & ViT-B/16  & $4.025$M (4.5\%)  & $4.910$M (5.5\%) \\
DINOv3  & ViT-L/16  & $4.157$M (1.4\%)  & $6.516$M (2.1\%) \\
\midrule
OpenCLIP & ViT-L/14 & $4.157$M (1.3\%)  & $6.516$M (2.1\%) \\
OpenCLIP & ViT-H/14 & $4.288$M (0.7\%)  & $8.220$M (1.3\%) \\
\bottomrule
\end{tabular}}
\caption{\textbf{Trainable parameter counts under various lightweight adaptation strategies.}
We report the number of trainable parameters when training only the segmentation head (frozen backbone, head-only) versus enabling LoRA-PEFT in the backbone (LoRA+head, rank $r=16$, $\alpha=32$, applied to attention \texttt{qkv} and \texttt{proj}). 
Percentages in parentheses indicate the fraction of trainable parameters relative to the total number of parameters in the corresponding end-to-end model.
The number of trainable parameters does not depend on the dataset considered for a fixed backbone and output label configuration.}
\label{tab:supp_trainable_params_all}
\end{table}

Notably, the reported trainable-parameter counts are identical for the matched DINOv2 and DINOv3 variants under our configuration. 
This is expected as the trainable budget in our setup is determined solely by the adaptation modules, namely the segmentation head and, when enabled, the LoRA adapters, whose parameterization depends on the backbone feature dimensionality (and on the set of adapted linear layers) rather than on the input image size. 
Since we apply the same LoRA configuration (rank $r{=}16$, $\alpha{=}32$, targets \texttt{attn.qkv} and \texttt{attn.proj}) and the same head architecture to the DINOv2 and DINOv3 variants with matching embedding dimensions, the resulting number of trainable parameters coincides.

\subsection{Evaluation metrics}\label{supp:metrics}

\paragraph{Fréchet DINOv2 Distance}
To quantitatively assess the mismatch between different data distributions, we rely on the Fréchet DINOv2 Distance (FD-DINOv2,~\cite{fdd}), calculated as follows. 
Given two sets of pooled embeddings $E_1$ and $E_2$, we approximate each $E_i$ by a multivariate Gaussian $\mathcal{N}(\mu_i,\Sigma_i)$ with empirical mean $\mu_i$ and covariance $\Sigma_i$, and compute
\begin{align}\label{eq:fddino}
\mathrm{FD\text{-}DINOv2} (E_1, E_2)
=
\lVert \mu_1 - \mu_2 \rVert_2^2
+
\mathrm{Tr}\!\left(\Sigma_1 + \Sigma_2 - 2(\Sigma_1^{1/2}\Sigma_2\Sigma_1^{1/2})^{1/2}\right).
\end{align}

\paragraph{Foreground IoU}
To quantitatively evaluate segmentation performance, we use the foreground IoU, calculated as follows.
With $p$ and $m$ denoting the predicted and ground-truth label maps, respectively, and considering the foreground indicators
$p_{\mathrm{fg}} = \mathbb{1}[p>0]$ and $m_{\mathrm{fg}} = \mathbb{1}[m>0]$, the foreground IoU is defined as
\begin{align}
\mathrm{IoU}_\mathrm{fg}
=
\frac{|\;p_{\mathrm{fg}} \cap m_{\mathrm{fg}}\;|}
     {|\;p_{\mathrm{fg}} \cup m_{\mathrm{fg}}\;| + \epsilon},
\end{align}
where $\epsilon$ is a small constant included for numerical stability ($1e-7$ in our experiments). 
Ground-truth masks can optionally be binarized at load time using a fixed threshold of $128$, corresponding to the midpoint for 8-bit $[0,255]$ masks (\emph{i.e.}, values $\ge 128$ are mapped to foreground and $<128$ to background), which affects $m$ (and therefore $m_{\mathrm{fg}}$) in the corresponding experiments. 
Model predictions are obtained by taking an \texttt{argmax} over the per-pixel logits.

\subsection{Effect of the unbalanced nature of the VNC dataset on paired training}\label{supp:unbalanced}

The paired training setting is inherently imbalanced as VNC provides substantially fewer labelled images than Lucchi++. 
To test whether this imbalance affects segmentation performance under paired training, we run an additional ablation with the frozen-backbone, head-only DINOv2-L/14 configuration. 
We compare our default unbalanced approach (standard concatenation and shuffling) against a balanced strategy that enforces $1+1$ batching.
This ensures that each mini-batch contains one Lucchi++ sample and one VNC sample, effectively upsampling the smaller dataset. 
In Table~\ref{tab:paired_balance_ablation}, we report IoU$_\mathrm{fg}$ separately on each test set as well as their macro-average.

\begin{table}[t]
\centering
\small
\setlength{\tabcolsep}{8pt}
\renewcommand{\arraystretch}{1.15}
\begin{tabular}{lccc}
\toprule
\textbf{Sampling} &  \textbf{Lucchi++ IoU$_\mathrm{fg}$} & \textbf{VNC IoU$_\mathrm{fg}$} & \textbf{Macro IoU$_\mathrm{fg}$} \\
\midrule
Unbalanced &  $0.720 \pm 0.004$ & $0.715 \pm 0.004$ & $0.717 \pm 0.004$ \\
Balanced & $\mathbf{0.0697 \pm 0.005}$ & $\mathbf{0.725 \pm 0.005}$ & $\mathbf{0.711 \pm 0.003}$ \\
\bottomrule
\end{tabular}
\caption{\textbf{Effect of dataset balancing on paired training.} 
Foreground IoU (IoU$_\mathrm{fg}$) segmentation performance of a frozen-backbone DINOv2-L/14 with head-only training in the paired dataset (Lucchi++ $\cup$ VNC) training regime reported as the mean$\pm$std over repeated runs.  
The balanced setting implements $1+1$ batching to ensure that each mini-batch contains one sample per dataset.}
\label{tab:paired_balance_ablation}
\end{table}

Our results indicate that balancing the datasets does not improve segmentation performance under paired training, with comparable macro IoU$_\mathrm{fg}$ values for balanced and unbalanced sampling.
While balancing slightly improves performance on VNC, it reduces performance on Lucchi++, resulting in a marginally lower macro IoU$_\mathrm{fg}$.
This suggests that, despite the substantial difference in the number of available images, the default unbalanced sampling strategy is sufficient to maintain segmentation performance across the two EM datasets.

\subsection{Domain mismatch diagnostics on unlabelled VNC Stack2}\label{supp:vncstack2}
In Sections~\ref{sec:res1} and~\ref{sec:res2}, we quantify representation-space domain mismatch between Lucchi++ and VNC using the labelled VNC split (Stack1), following the same supervised setting used throughout the segmentation experiments. 
However, this labelled test subset contains only $3$ slices, which makes distribution-level diagnostics such as feature-space distances potentially sensitive to small-sample effects. 
Since our domain-mismatch analysis is fully unsupervised and does not require masks, we repeated the experiment on the unlabelled VNC Stack2, which provides a larger set of $20$ images for evaluation.

\begin{figure}[t]
    \centering
    \begin{subfigure}[t]{0.49\linewidth}
        \centering
        \includegraphics[width=\linewidth]{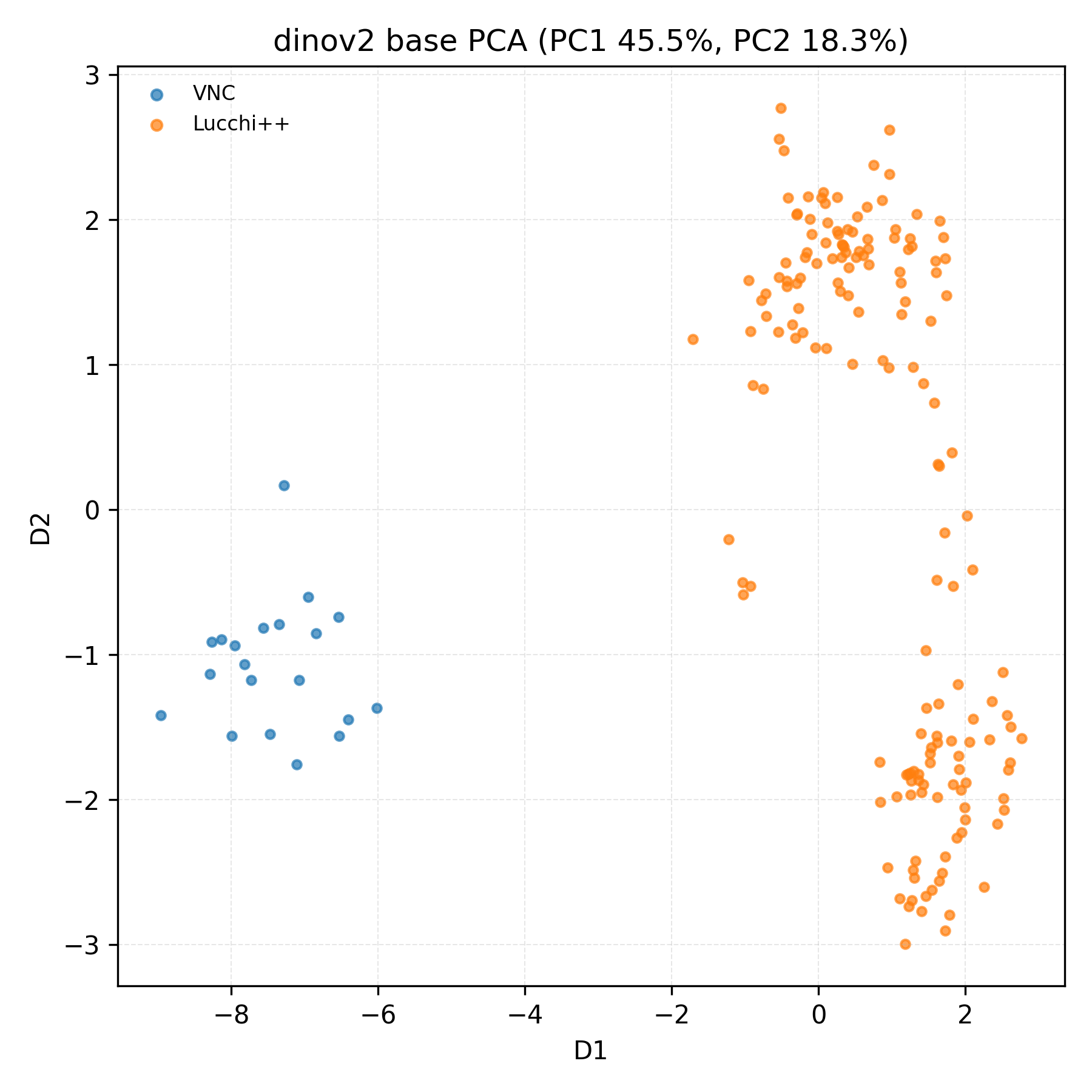}
        \caption{Frozen backbone (head-only training).}
        \label{fig:pca_stack2_a}
    \end{subfigure}
    \hfill
    \begin{subfigure}[t]{0.49\linewidth}
        \centering
        \includegraphics[width=\linewidth]{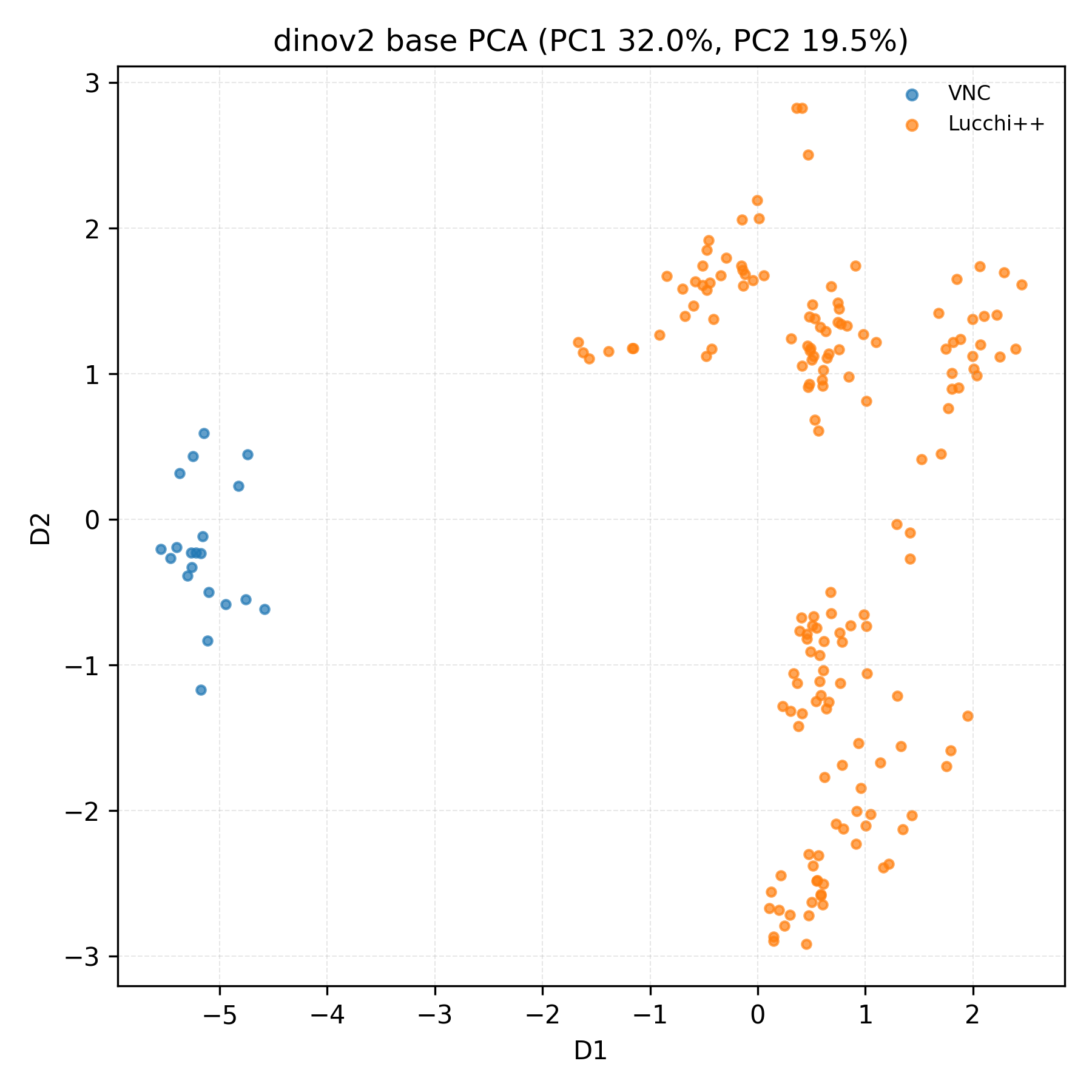}
        \caption{LoRA-PEFT backbone.}
        \label{fig:pca_stack2_b}
    \end{subfigure}

    \caption{\textbf{Domain mismatch between Lucchi++ and VNC (Stack2) in DINOv2 feature space.} 
    (a) PCA projection of image-level embeddings from the frozen DINOv2-L/14 backbone exhibits no overlap between the two datasets.
    (b) With LoRA enabled, the feature-space gap is minimally reduced, and the two domains remain clearly separable.}
    \label{fig:pca_stack2_both}
\end{figure}

Concretely, we extracted one image-level embedding per slice from DINOv2 by global pooling of the final-layer patch tokens, as in Figure~\ref{fig:pca_mismatch_both}, and then carried out PCA on the joint set of Lucchi++ test images and VNC Stack2 slices. 
In Figure~\ref{fig:pca_stack2_both}, we visualize the first two principal components in the frozen backbone (head-only training) and after applying LoRA-PEFT.
To complement this visualization, we also report the FD-DINOv2 (\eqref{eq:fddino}) and the linear separability of the domain label (Lucchi++ vs.\ VNC) using an LR probe.

We observe that data points from Lucchi++ and the VNC Stack2 remain clearly separated in feature space, which is consistent with the conclusions reached considering the labelled VNC split. 
Quantitatively, the FD-DINOv2 remains large for the frozen backbone and decreases when LoRA is applied ($82.06$ and $43.26$, respectively), but the two datasets are still perfectly linearly separable (accuracy/AUROC = $1.0$ in both cases). 
Overall, increasing the number of VNC images from a handful of labelled test slices to the full unlabelled Stack2 therefore does not alter the qualitative outcome: the domain mismatch between the two datasets persists and is only marginally reduced by LoRA, despite the comparable segmentation performance achieved under paired and single-dataset training.

\clearpage

\begingroup

\thispagestyle{empty} 

\vspace*{1cm}

\begin{center}
    {\Large\bfseries
    Corrigendum to the paper:\\[0.3em]
    Are Vision Foundation Models Foundational for Electron Microscopy Image Segmentation?\par}

    \vspace{0.8cm}

    {\normalsize
    Caterina Fuster-Barceló$^{1}$, Virginie Uhlmann$^{1}$\par}

    \vspace{0.4cm}

    {\small
    $^{1}$Department of Molecular Life Sciences,\\ Universität Zürich, Zürich, Switzerland}\\
\end{center}

\vspace{1em}

An earlier version of this paper, presented at the Learning Meaningful Representations of Life (LMRL) workshop at ICLR 2026, reported that jointly training vision foundation models (VFMs) on heterogeneous electron microscopy (EM) datasets led to a substantial decrease in segmentation performance compared with training on individual datasets. We also observed that the datasets remained clearly separable in the models' feature space, even after parameter-efficient fine-tuning (PEFT). These findings suggested that domain differences could limit the effectiveness of joint training.

We however later identified an error in the code we used for the joint-training experiments. After correcting it, we reran the experiments and no longer observed the previously reported performance degradation. We thus revised the manuscript and its conclusions accordingly.

The corrected results show that joint training generally maintains or improves segmentation performance compared with training on individual datasets. Interestingly, the domain separability previously observed in the feature space remains, including after PEFT. These findings suggest that feature-space separability does not necessarily translate into poor downstream segmentation performance.

Although these new results change our original hypothesis, they provide equally interesting insights into the behaviour of VFMs across heterogeneous microscopy datasets, which can seemingly successfully learn from distinct domains even when their representations remain separable. This highlights the importance of evaluating downstream performance alongside feature-space analyses when assessing model transferability.

\clearpage
\endgroup

\end{document}

%% file: math_commands.tex
\usepackage{amsmath,amsfonts,bm}

\def\eqref#1{equation~\ref{#1}}

\def\1{\bm{1}}

\DeclareMathAlphabet{\mathsfit}{\encodingdefault}{\sfdefault}{m}{sl}
\SetMathAlphabet{\mathsfit}{bold}{\encodingdefault}{\sfdefault}{bx}{n}

